\documentclass[12pt,a4paper]{article}

\pdfoutput=1

\usepackage[utf8]{inputenc}
\usepackage[english]{babel}

\usepackage{hyperref}

\usepackage{indentfirst}

\usepackage{graphicx}
\usepackage{subcaption}
\usepackage[section]{placeins}

\usepackage{bm}
\usepackage{amsmath}
\usepackage[ruled,vlined]{algorithm2e}

\usepackage[numbers]{natbib}
\title{A Novel VHR Image Change Detection Algorithm Based on Image Fusion and Fuzzy C-Means Clustering}

\date{}

\usepackage{authblk}
\author[1]{Rongcui Dong}
\author[2,3]{Haoxiang Wang}
\affil[1]{Viterbi School of Engineering \\
University of Southern California \\
LA, USA}
\affil[2]{Department of ECE \\
Cornell University \\
NY, USA}
\affil[3]{
GoPerception Laboratory \\
NY, USA}

\begin{document}

\maketitle

\begin{abstract}

This thesis describes a study to perform change detection on Very High Resolution satellite images using image fusion based on 2D Discrete Wavelet Transform and Fuzzy C-Means clustering algorithm. Multiple other methods are also quantitatively and qualitatively compared in this study.

\end{abstract}

\section{Introduction}\label{sec:introduction}

Change detection, an automated process to detect difference between two temporally separated image, have important applications in many fields. When applied to remote sensing, such as images obtained from satellites, change detection can be employed research global or local ecology \cite{kerr2003space} \cite{coppin1996digital}, urban and land cover changes \cite{yang2003urban} \cite{weng2001remote}, disaster detection and assessment \cite{stramondo2006satellite} \cite{tralli2005satellite}, \emph{etc.}. As current satellite imaging technology enters sub-meter definition, the amount of data needed to processed grows rapidly, and older algorithms suitable for lower resolution may not work against these much more detailed images \cite{hussain2013change}. Methods to work with VHS images include pixel-based and object-based algorithms, where the former works directly on individual pixels while the latter groups multiple pixels into objects. Pixel-based methods are generally easier to understand and interpret, but lacks the ability to consider spatial context \cite{hussain2013change}. Usually, some type of \emph{difference map} representing the difference between two images are used, and the algorithm outputs a binary \emph{change map} where changed area is marked. This study attempts to use 2D Discrete Wavelet Transform to combine advantages of several types of difference map, and use Fuzzy C-Means clustering to create a binary change map.

\section{Difference Map Fusion with Discrete Wavelet Transform}
\label{sec:dwt}

There are many methods to obtain the difference between two images. Of the two most simple ones are pixel-by-pixel subtraction and division. In this study, the former is called \emph{minus map}, and the latter is called \emph{ratio map}. The resulting images are combined in frequency domain of 2-D discrete wavelet transform.

2-D discrete wavelet transform maps images from spatial domain to frequency domain. To perform a 2DDWT on an image, a 1-Dimensional DWT is performed on each row (or column) of pixels, and then another 1DDWT is performed on each column (or row) on the partially transformed image \cite{radley2016green} \cite{niu2010multi}. In the proposed method, Haar transform is used due to its relative simplicity. Wavelet transform produces multiple sub-images, where the upper-left portion is the high frequency component, and each band extending outwards represents a lower frequency component \cite{han2010multi}. In this study, the algorithm to perform Haar transform is derived from its matrix form \cite{porwik2004haar}, as shown in Algorithm \ref{alg:haar}, where $\bm{X}$ is the square input image with edge length padded to the nearest $L=2^n$, and $\bm{H}$ is a 1D Haar transform matrix which transforms each columns of $\bm{X}$ individually.

\begin{algorithm}[htb]
\SetAlgoLined
\KwData{$\bm{X}$}
\KwResult{$\bm{\hat{X}}$}
$\bm{\hat{X}} \gets {1\over L}\bm{H}(\bm{H}\bm{X}^T)^T$
\caption{Haar transform}
\label{alg:haar}
\end{algorithm}

\begin{figure*}[htb]
  \centering
  \begin{subfigure}[b]{0.2\textwidth}
    \includegraphics[width=\textwidth]{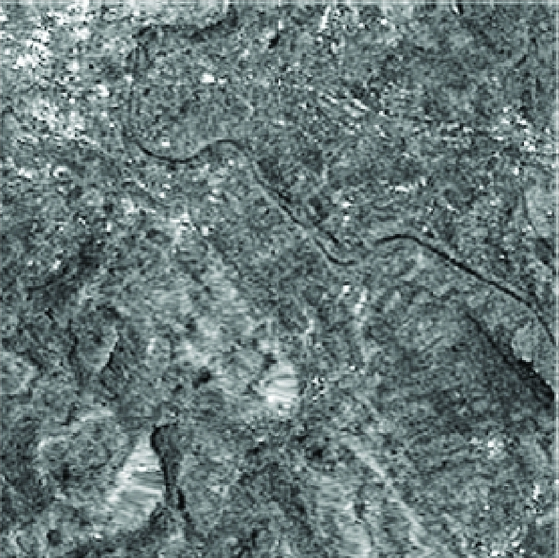}
  \end{subfigure}
  \begin{subfigure}[b]{0.2\textwidth}
    \includegraphics[width=\textwidth]{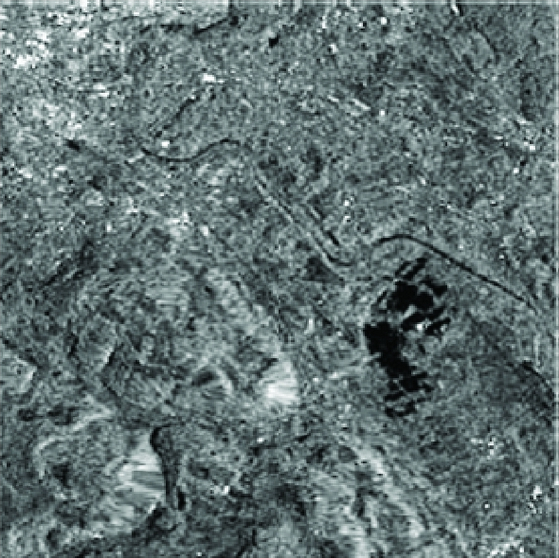}
  \end{subfigure}
  \begin{subfigure}[b]{0.2\textwidth}
    \includegraphics[width=\textwidth]{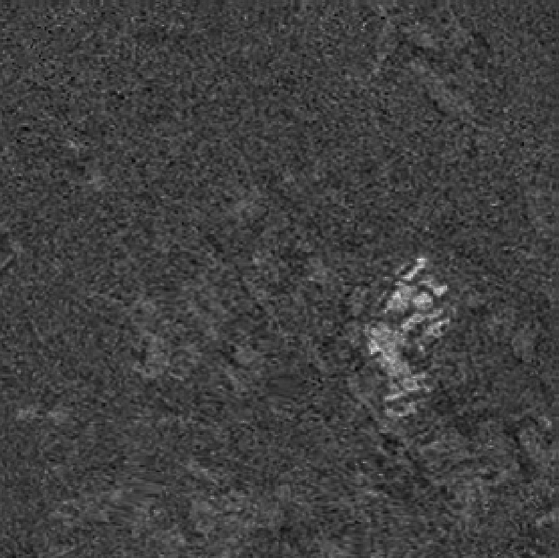}
  \end{subfigure}
  \begin{subfigure}[b]{0.2\textwidth}
    \includegraphics[width=\textwidth]{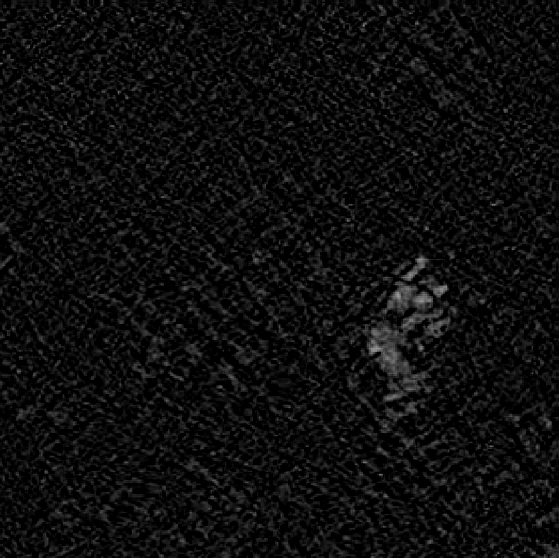}
  \end{subfigure}

  \begin{subfigure}[b]{0.2\textwidth}
    \includegraphics[width=\textwidth]{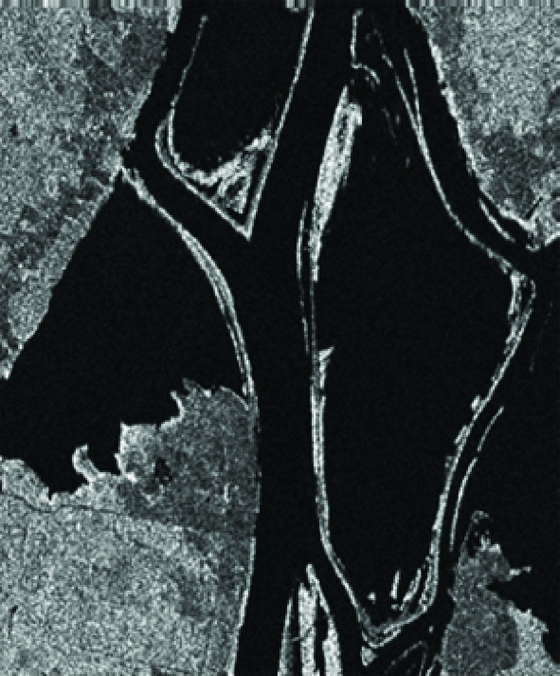}
  \end{subfigure}
  \begin{subfigure}[b]{0.2\textwidth}
    \includegraphics[width=\textwidth]{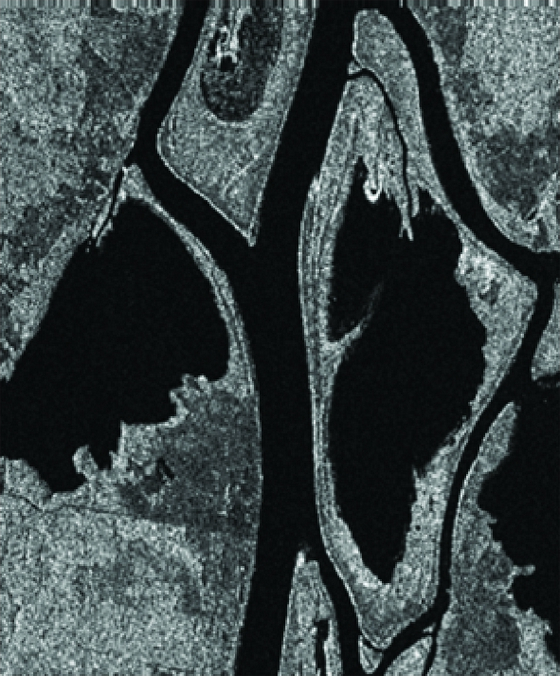}
  \end{subfigure}
  \begin{subfigure}[b]{0.2\textwidth}
    \includegraphics[width=\textwidth]{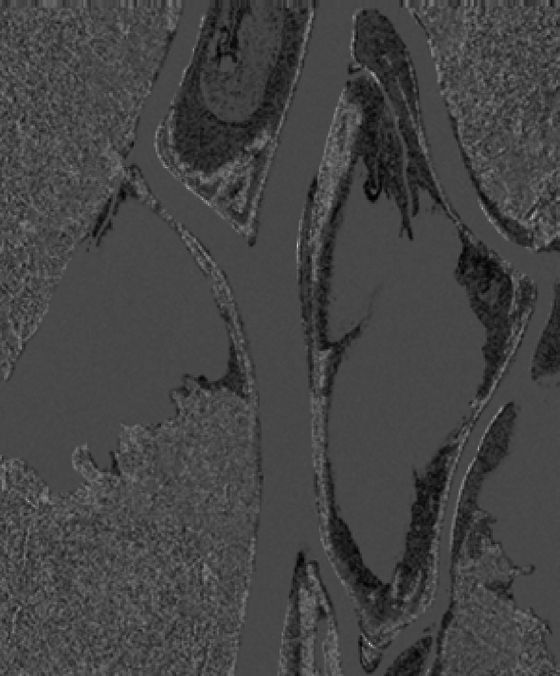}
  \end{subfigure}
  \begin{subfigure}[b]{0.2\textwidth}
    \includegraphics[width=\textwidth]{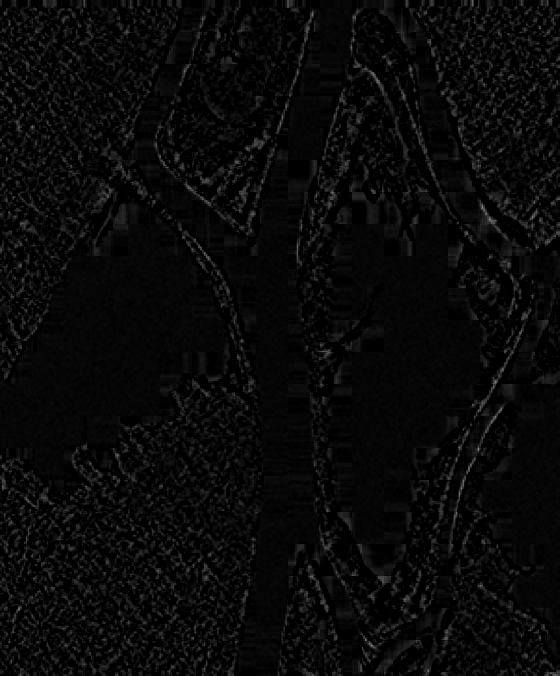}
  \end{subfigure}

  \begin{subfigure}[b]{0.2\textwidth}
    \includegraphics[width=\textwidth]{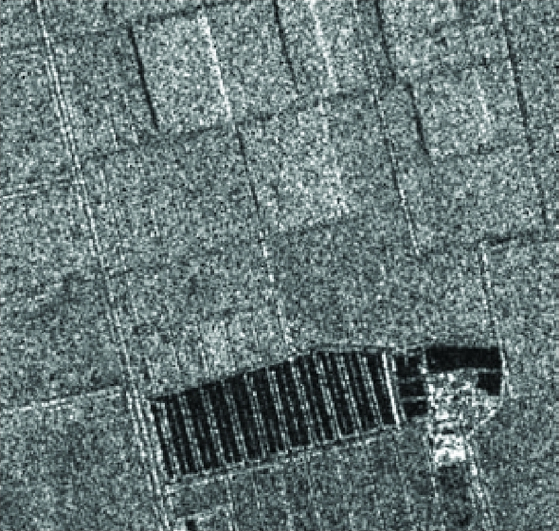}
  \end{subfigure}
  \begin{subfigure}[b]{0.2\textwidth}
    \includegraphics[width=\textwidth]{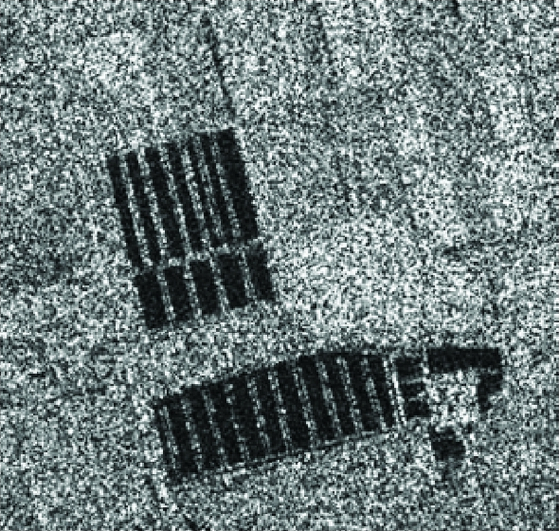}
  \end{subfigure}
  \begin{subfigure}[b]{0.2\textwidth}
    \includegraphics[width=\textwidth]{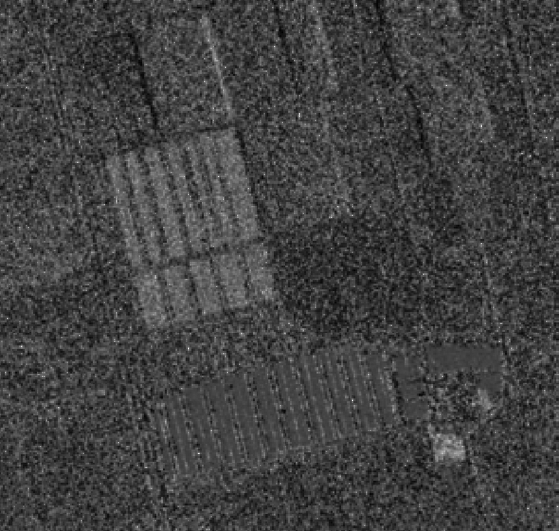}
  \end{subfigure}
  \begin{subfigure}[b]{0.2\textwidth}
    \includegraphics[width=\textwidth]{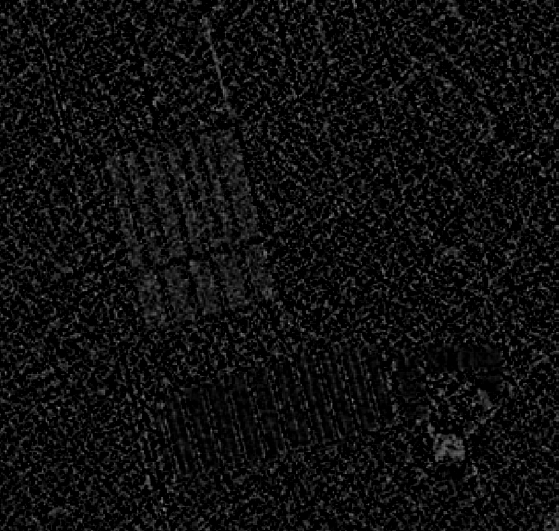}
  \end{subfigure}
  \caption{Fused Differential Maps. Left two columns are source images; third column are fused by weighted average; fourth column are fused by wavelet transform}
  \label{fig:fused-diff-map}
\end{figure*}

In order to combine the minus map and ratio map into a single differential map that can be used by segmentation algorithm, the frequency separation property of the 2DDWT is exploited. In this case, the goal is to combine the advantages of minus map and ratio map. Therefore, the lower half frequency components of the minus map and higher half frequency components of the ratio map are combined into a new frequency domain image. An inverse 2DDWT is performed on the resulting image to obtain fused difference map. The output of the image fusion is shown in Figure \ref{fig:fused-diff-map}. As a comparison, the simpler weighted average method is also shown. It should be noticed that 2DDWT fusion gives much cleaner, but dimmer images. This lack of contrast can be a source of problem encountered in the later segmentation stage.

\section{Fuzzy C-Means Clustering}

The result of a fused difference map usually contain large amount of disconnected sectors. In order to generate a usable binary change map, clustering algorithm can be used to identify relavent groups of pixels. Fuzzy C-Means (FCM) clustering algorithm is widely used in remote imaging change detection because it ``retains more information from the original image and has robust characteristics for ambiguity'' \cite{gong2012change}. However, the original FCM is succeptable to noise since it does not use any spatial information \cite{gong2012change}.

In this experiment, although the original FCM is claimed to be noisy, it is used nevertheless, due to its relative simplicity. Algorithm \ref{alg:fcm} shows the pseudo code of the FCM algorithm used in the experiment, where vector $\bm{x}$ is the difference map flattened into a vector, matrix $\bm{U}$ is the membership matrix of size $N \times c$ where $c$ is the number of clusters and $N$ is the length of $\vec{x}$, vector $\bm{v}$ is the centers of clusters with size $c$, $m$ is a parameter determining the degree of fuzzyness, and any symbols with hats like $\hat{\bm{U}}$ or $\hat{\bm{v}}$ are the computed new value in each iteration.. At initialization, $\bm{U}$ is random initialized with elements between $0$ and $1$. The results are the final values of membership matrix $\bm{U}$ and center vector $\bm{v}$. In this experiment, the cluster which a pixel $i$ belongs to is determined by the highest score of its corresponding membership vector $\bm{U}_i$ of size $c$. Since the change map is binary, the cluster with highest center is treated as \emph{changed} and others are \emph{unchanged}. In this experiment, $c$ is set to $6$ and $q$ set to $2$ arbitrarily.

\begin{algorithm}[htb]
\SetAlgoLined
\SetKwRepeat{Do}{do}{while}
\KwData{$\bm{x}$, $\bm{U}$}
\KwResult{Final values of $(\bm{U}, \bm{v})$}
$\bm{x} \gets$ Flattened difference map\;
$\bm{U} \gets \bm{U_{rand(0,1)}}$\;
\Do{$\lVert \bm{U} - \hat{\bm{U}} \rVert_{L_2} > \epsilon$}{
\For{$j$ $\rm{in}$ $1..c$}{
$\hat{\bm{v}}_j \gets \left(\frac{\sum^N_{i=1}u^m_{ij}\cdot x_i}{\sum^N_{i=1}u^m_{ij}}\right)$\;
}
\For{$i$ $\rm{in}$ $1..N$} {
\For{$j$ $\rm{in}$ $1..c$} {
$\hat{\bm{U}}_{ij} \gets \left({1 \over \sum^C_{k=1}\left( \frac{\lVert \bm{x}_i-\hat{\bm{v}}_j \rVert}{\lVert \bm{x}_i-\hat{\bm{v}}_k \rVert} \right)^{2/(m-1)} }\right)$\;
}
}
$\bm{U} \gets \hat{\bm{U}}$\;
$\bm{v} \gets \hat{\bm{v}}$\;
}
\caption{Fuzzy C-Means algorithm}
\label{alg:fcm}
\end{algorithm}

\section{Experimental Study}
\label{sec:results}

\begin{figure*}[htb]
  \centering
  \begin{subfigure}[b]{0.3\textwidth}
    \includegraphics[width=\textwidth]{imgs/ground-truth/B1}
  \end{subfigure}
  \begin{subfigure}[b]{0.3\textwidth}
    \includegraphics[width=\textwidth]{imgs/ground-truth/B2}
  \end{subfigure}
  \begin{subfigure}[b]{0.3\textwidth}
    \includegraphics[width=\textwidth]{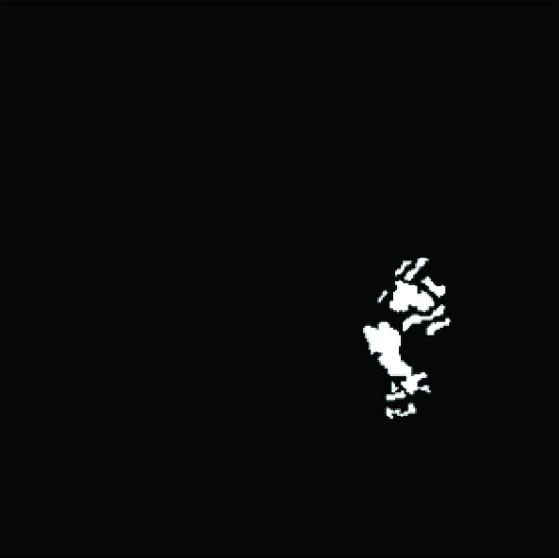}
  \end{subfigure}

  \begin{subfigure}[b]{0.3\textwidth}
    \includegraphics[width=\textwidth]{imgs/ground-truth/O1}
  \end{subfigure}
  \begin{subfigure}[b]{0.3\textwidth}
    \includegraphics[width=\textwidth]{imgs/ground-truth/O2}
  \end{subfigure}
  \begin{subfigure}[b]{0.3\textwidth}
    \includegraphics[width=\textwidth]{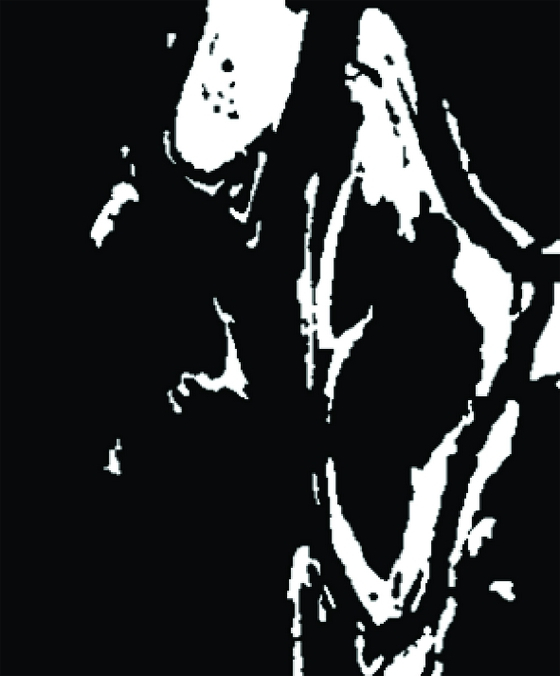}
  \end{subfigure}

  \begin{subfigure}[b]{0.3\textwidth}
    \includegraphics[width=\textwidth]{imgs/ground-truth/Y1}
  \end{subfigure}
  \begin{subfigure}[b]{0.3\textwidth}
    \includegraphics[width=\textwidth]{imgs/ground-truth/Y2}
  \end{subfigure}
  \begin{subfigure}[b]{0.3\textwidth}
    \includegraphics[width=\textwidth]{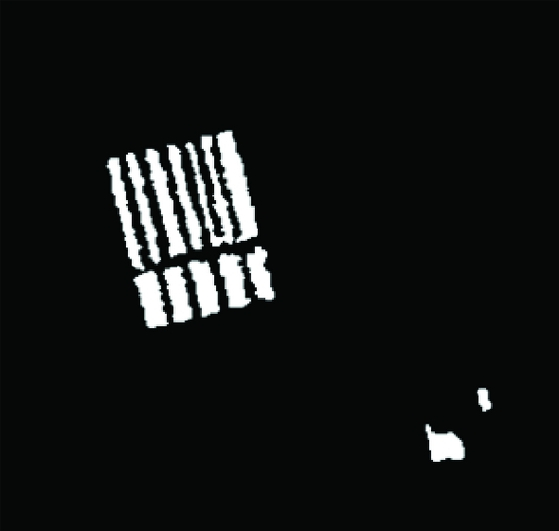}
  \end{subfigure}
  \caption{Test and Ground Truth Images.}
  \label{fig:test-and-ground-truth}
\end{figure*}

In this study, three other segmentors are also implemented to compare with the proposed method: Otsu's threshold method \cite{otsu79}, K-Means clustering method \cite{macqueen1967some}, and a Superpixel and neural network based algorithm \cite{gong2017superpixel}.In order to reliably compare different methods, change maps are compared pixel-by-pixel with corresponding ground truth images (as shown in Figure \ref{fig:test-and-ground-truth}), and true positive, true negative, false positive, false negative, and Cohen's Kappa are calculated. The closer true positive, true negative, and kappa are to $1$, the more accurate a detector performs.

\begin{figure*}[htb]
  \centering
  \begin{subfigure}[b]{0.19\textwidth}
    \includegraphics[width=\textwidth]{imgs/ground-truth/B0}    
  \end{subfigure}
  \begin{subfigure}[b]{0.19\textwidth}
    \includegraphics[width=\textwidth]{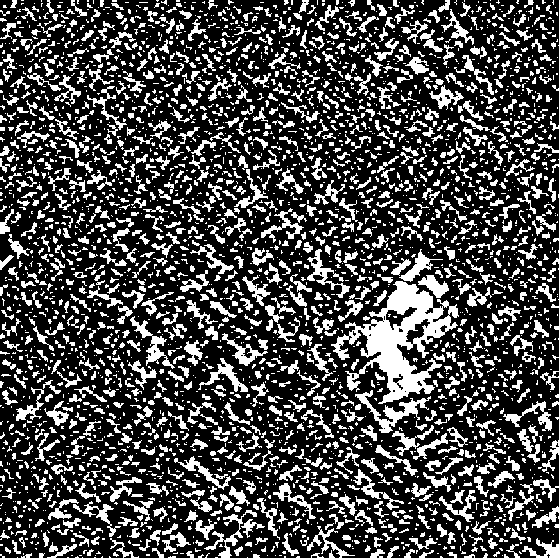}    
  \end{subfigure}
  \begin{subfigure}[b]{0.19\textwidth}
    \includegraphics[width=\textwidth]{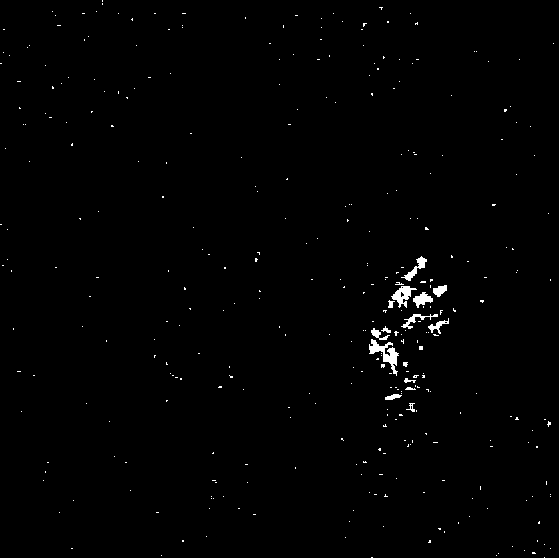}    
  \end{subfigure}
  \begin{subfigure}[b]{0.19\textwidth}
    \includegraphics[width=\textwidth]{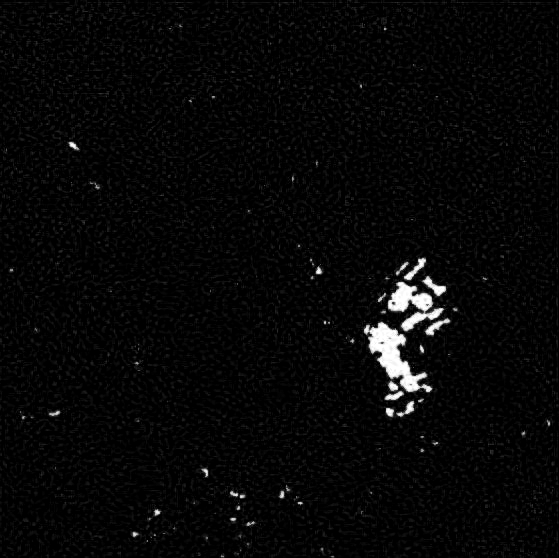}    
  \end{subfigure}
  \begin{subfigure}[b]{0.19\textwidth}
    \includegraphics[width=\textwidth]{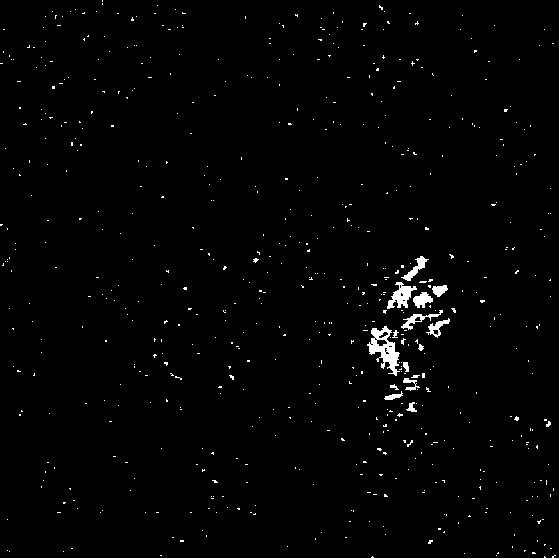}    
  \end{subfigure}

  \begin{subfigure}[b]{0.19\textwidth}
    \includegraphics[width=\textwidth]{imgs/ground-truth/O0}    
  \end{subfigure}
  \begin{subfigure}[b]{0.19\textwidth}
    \includegraphics[width=\textwidth]{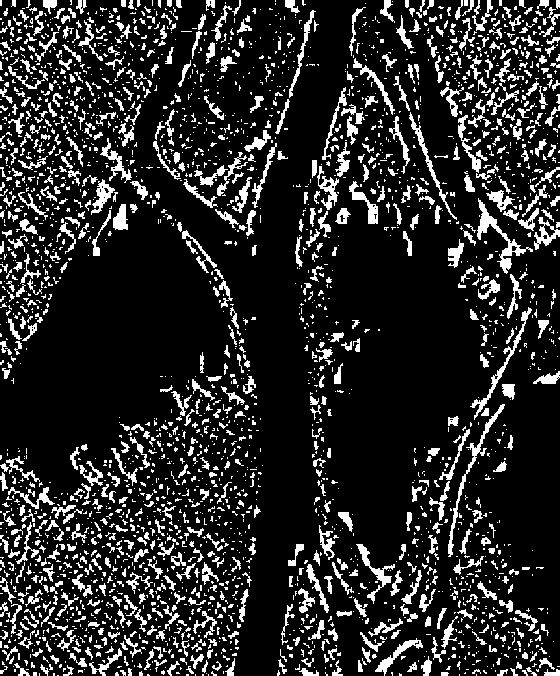}    
  \end{subfigure}
  \begin{subfigure}[b]{0.19\textwidth}
    \includegraphics[width=\textwidth]{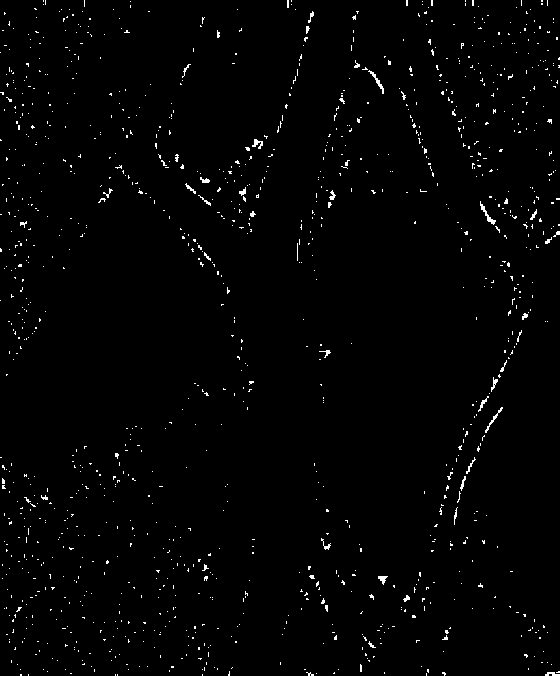}    
  \end{subfigure}
  \begin{subfigure}[b]{0.19\textwidth}
    \includegraphics[width=\textwidth]{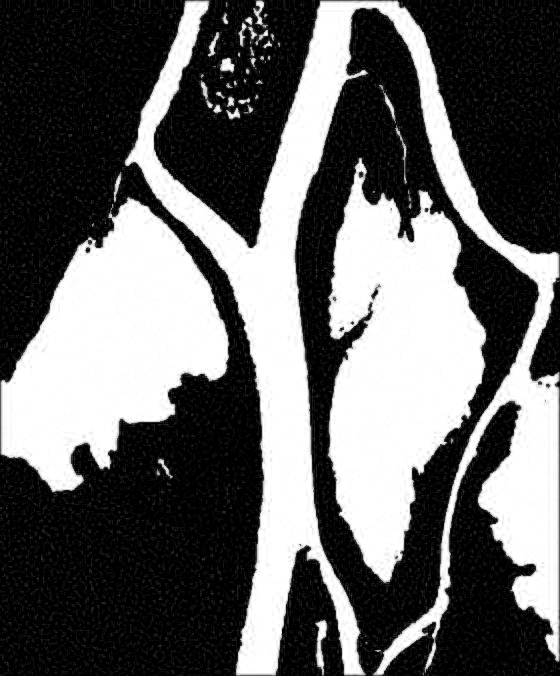}    
  \end{subfigure}
  \begin{subfigure}[b]{0.19\textwidth}
    \includegraphics[width=\textwidth]{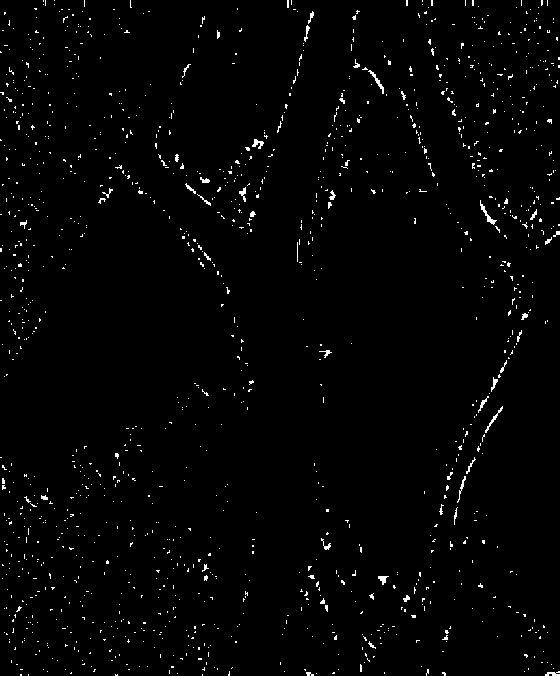}    
  \end{subfigure}

  \begin{subfigure}[b]{0.19\textwidth}
    \includegraphics[width=\textwidth]{imgs/ground-truth/Y0}    
  \end{subfigure}
  \begin{subfigure}[b]{0.19\textwidth}
    \includegraphics[width=\textwidth]{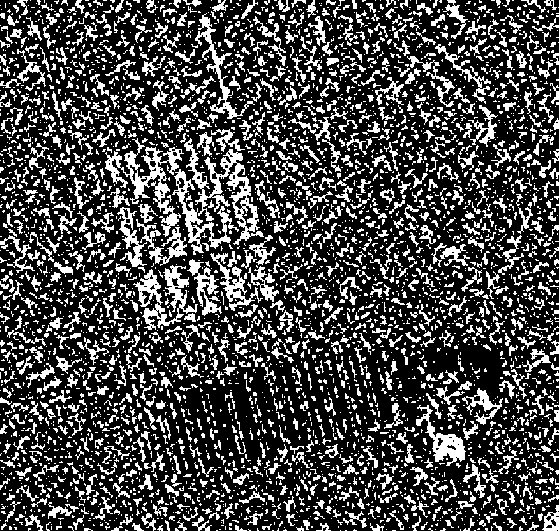}    
  \end{subfigure}
  \begin{subfigure}[b]{0.19\textwidth}
    \includegraphics[width=\textwidth]{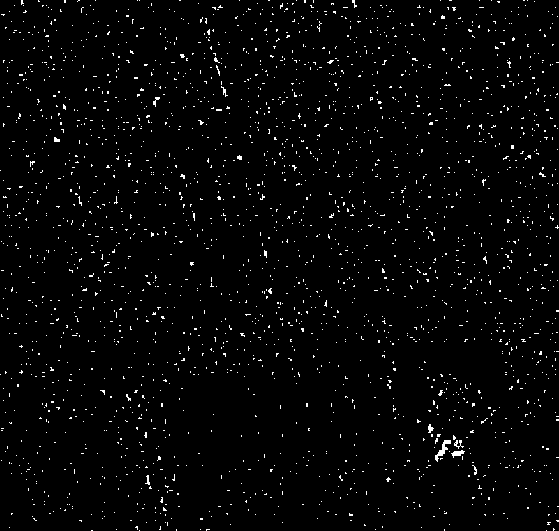}    
  \end{subfigure}
  \begin{subfigure}[b]{0.19\textwidth}
    \includegraphics[width=\textwidth]{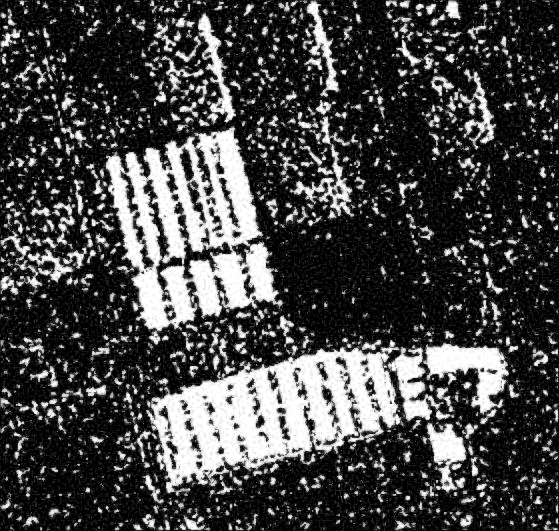}    
  \end{subfigure}
  \begin{subfigure}[b]{0.19\textwidth}
    \includegraphics[width=\textwidth]{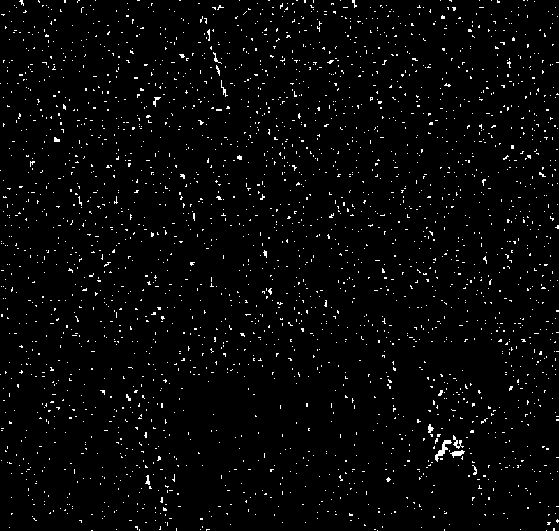}    
  \end{subfigure}
  \caption{Results. The leftmost row is ground truth, and to the right (in order) are Otsu's method, K-Means clustering, SuperPixel based, and Fuzzy C-Means clustering (the proposed method).}
  \label{fig:results}
\end{figure*}

The segmentation results are shown in Figure \ref{fig:results} and Table \ref{tab:results}. As shown in Table \ref{tab:results}, and possibly visible from the figures, the first set of test images are the easiest to segment for each segmentor. The second set, however, causes confusion to all methods, while the third is not much better. 

In this experiment, the Otsu's threshold method is expected to fail since a simple binary threshold does not take local spatial information into account at all, and therefore is very sensitive to spot noise, as illustrated both by the noisy image and high false positive. K-means and FCM has comparable performance, but neither does very well on the second and third set. The similarity in performance is probably due to the small fuzzyness used in this experiment.

It is not surprising that the Superpixel based method achieves better accuracy, although it almost precisely misses the changed part in test set two, as demonstrated by its false negative coming near 1. In contrast to other methods, the Superpixel method is the only one that is object-based \cite{gong2017superpixel} instead of pixel-based, and employs machine learning to refine the results, and therefore accounts for contextual information much better.

\begin{table}[htb]
\centering
\begin{tabular}{|c|c|c|c|c|c|c|}
  \hline
  Method & Test Set & TP & FP & TN & FN & $\kappa$ \\\hline\hline
  Otsu & 1 &0.9697&0.2737&0.7263&0.0303&0.0601 \\\hline
  Otsu & 2 &0.1173&0.1677&0.8323&0.8827&-0.0503 \\\hline
  Otsu & 3 &0.6369&0.2613&0.7387&0.3631&0.1182 \\\hline\hline
  KMeans &1&0.4253&0.0028&0.9972&0.5747&0.5131 \\\hline
  KMeans &2&0.0041&0.0156&0.9844&0.9959&-0.0183 \\\hline
  KMeans &3&0.0507&0.0205&0.9795&0.9493&0.0410 \\\hline\hline
  SuperPixel & 1 & 0.8307 & 0.0027 & 0.9973 & 0.1693 & 0.8119 \\\hline
  SuperPixel & 2 & 0.0474 & 0.4492 & 0.5508 & 0.9526 & -0.2541 \\\hline
  SuperPixel & 3 & 0.9830 & 0.2100 & 0.7900 & 0.0170 & 0.2684 \\\hline\hline
  FCM & 1 & 0.5602 & 0.0063 & 0.9937 & 0.4398 & 0.5400 \\\hline
  FCM & 2 & 0.0049 & 0.0183 & 0.9817 & 0.9951 & -0.0210 \\\hline
  FCM & 3 & 0.0585 & 0.0239 & 0.9761 & 0.9415 & 0.0449 \\\hline
\end{tabular}
  \caption{Results.}
  \label{tab:results}
\end{table}

\section{Conclusion}
\label{sec:conclusion}

In summary, the method explored in this study did not yield optimal results. The problem likely originates from the noisy differential maps, as the changed area is not significantly brighter than the unchanged ones. In further study, other types of differential maps may be tested, and denoising algorithm can be employed before performing segmentation.

\newpage

\bibliography{Citations}

\end{document}